\documentclass[11pt]{article}

\usepackage{amssymb, amsmath, epsfig, amsthm, enumerate}

\theoremstyle{plain}
  \newtheorem{thm}{Theorem}[section]
  
  \newtheorem{lemma}[thm]{Lemma}
  \newtheorem{cor}[thm]{Corollary}
\theoremstyle{definition}
  \newtheorem{defi}[thm]{Definition}

 \theoremstyle{remark}
  \newtheorem{rmk}[thm]{Remark}

\newcommand{\mn}{{\boldsymbol{n}}}

\newcommand{\mR}{\mathbb{R}}

\newcommand{\Om}{\Omega}
\newcommand{\PQ}{\partial Q}
\newcommand{\Gr}{{\Gamma_{\re}}}
\newcommand{\Gi}{{\Gamma_{\im}}}
\newcommand{\cC}{\mathcal{C}}
\newcommand{\mc}{\mathbf{c}}
\newcommand{\mt}{{\boldsymbol{t}}}
\newcommand{\ep}{\varepsilon}

\newcommand{\pL}{\partial L}

\newcommand{\pd}{\partial}
\newcommand{\union}{\cup}   %% set operations
\newcommand{\joint}{\cap}
\newcommand{\re}{{\mathrm{re}}}
\newcommand{\im}{{\mathrm{im}}}
\newcommand{\idle}{{\mathrm{idle}}}

\title{ First-Order Modeling and Stability Analysis of Illusory Contours}
\author{Yoon Mo Jung and Jianhong (Jackie) Shen
        \thanks{This work has been partially supported by the NSF (USA) under grant number
                DMS-0202565.  Mail address of the authors:
                School of Mathematics, University of Minnesota, 206 Church Street SE,
                Minneapolis, MN 55455, USA.
                Corresponding author: Jackie Shen, jhshen@math.umn.edu, (612) 625-3570 (Tel),
                www.math.umn.edu/\~{}jhshen.} }

\date{}

\begin{document}

\maketitle

\begin{center}{\em \small Dedicated to David Mumford
        -- A wellspring of inspiration to the young.}\end{center}

\begin{abstract}
In visual cognition, illusions help elucidate certain intriguing latent perceptual
functions of the human vision system, and their proper mathematical modeling and
computational simulation are therefore deeply beneficial to both biological and computer
vision. Inspired by existent prior works, the current paper proposes a first-order
energy-based model for analyzing and simulating illusory contours. The lower complexity
of the proposed model facilitates rigorous mathematical analysis on the detailed
geometric structures of illusory contours. After being asymptotically approximated by
classical active contours, the proposed model is then robustly computed using the
celebrated level-set method of Osher and Sethian ({\em J. Comput. Phys.}, {\bf 79}:12-49,
1988) with a natural supervising scheme. Potential cognitive implications of the
mathematical results are addressed, and generic computational
examples are demonstrated and discussed. \\[1ex]

\noindent {\bf Keywords:} illusion, illusory contours, energy model,  local stability,
contour decomposition, real and imaginary, kinks, spans, turns, level-set method,
supervision.

\end{abstract}

\section{Introduction}

In System Theory~\cite{oppwil}, input-output analysis has been a major tool for partial
or complete identification of black-box systems. In cognitive vision science, the study
of various visual illusions follows exactly the same spirit. Cognitive scientists have
designed numerous intriguing inputs of image signals, so that the distorted or
transformed outputs (as reported by an average human observer) can help reveal some
crucial latent properties of the human vision system (see, e.g., the remarkable works of
Adelson~\cite{ade00}, Knill and Kersten~\cite{ker,illu_KnillKersten}, and
Kanizsa~\cite{kan}). {\em Illusory contours} are such a well known class of visual
illusions, and the current paper develops a mathematical model to characterize, analyze,
and simulate generic illusory contours. Our work has been closely inspired by many
existent modeling works, especially by Sarti, Malladi, and Sethian~\cite{sarmalset}, and
Zhu and Chan~\cite{illu_zhuchan,illu_zhuchan_shape}.

\begin{figure}[ht]
 \centering{
 \epsfig{file=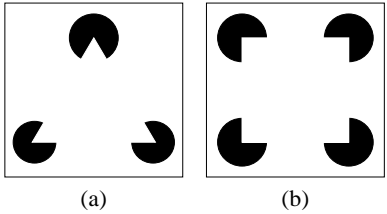, height=1.1in, width=2in}
 \caption{Two classical examples of illusory contours: Kanizsa's triangle and square.}
 \label{Fig11}  }
\end{figure}

Fig.~\ref{Fig11} shows two examples of illusory contours known as \textit{Kanizsa
triangle and square}~\cite{kan,sarmalset,illu_zhuchan}. The illusory or {\em imaginary}
white triangle and square pop out almost instantly to a normal observer. The human vision
system is capable of interpolating non-existent edges and making meaningful visual
organization of both the {\em real} and {\em imaginary} edge segments. From the
viewpoints of modeling and computation, such illusory perception has also been called
{\em segmentation with missing boundaries}~\cite{sarmalset}, and is closely related to
some earlier works~(e.g.,~\cite{mumsha,nitmumshi}).

\begin{figure}[ht]
 \centering{
 \epsfig{file=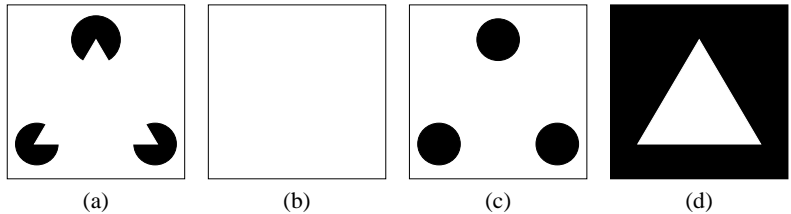, height=1.1in, width=4.1in}
 \caption{Multi-layer decomposition of Kanizsa's triangle: (a) image of Kanizsa's triangle,
          (b) far background, (c) three disks in the middle range, and (d) the illusory triangle
              closest to a viewer.}
 \label{Fig12}  }
\end{figure}

Qualitatively, the perception of illusory contours can be deciphered by the Gestalt
framework~\cite{fre94,kan}. In a typical example of illusory contours, the illusory
target objects share the same gray level as their background and become invisible to
ordinary detectors. The illusory targets however do leave their footprints, mainly
through the cues of occlusion, e.g., the visible corners in Fig.~\ref{Fig11}. An average
human vision system can organize these cues and develop a so-called 2.1-D sketch of the
scene~\cite{nitmumshi}, i.e., to separate and complete objects in ordered (according to
the relative depth to a viewer) multiple layers. The multi-layer sketch of Kanizsa's
triangle is displayed in Fig.~\ref{Fig12}, for example. We also refer the reader to the
recent work of Shen~\cite{she_monoid} on an abstract study of the occlusion phenomenon
and 2.1-D models.

Quantitatively, however, it is a nontrivial task to properly model and compute illusory
contours. There are two notable recent works that have primarily influenced the present
work in developing plausible mathematical models and robust computational schemes.

In~\cite{sarmalset}, Sarti, Malladi, and Sethian first proposed a variational-PDE model
for computing illusory contours based on a \textit{reference point} within an image.
Given a reference point which models the focus of visual attention, a surface is then
constructed and flattened except along the existing {\em real} edges. The level sets of
the surface function are able to connect the imaginary  contours. Define the edge
indicator function by~\cite{sarmalset}
$$g(x)=g(x_1, x_2)=\frac{1}{1+(\lvert \nabla G_\sigma (x)*u(x)\rvert/\beta)^2}, $$
$$ G_\sigma(\xi)=\frac{\exp(-(\xi/\sigma)^2)}{\sigma\sqrt{\pi}}. $$
Then the model is to minimize the $g$-weighted area of the surface $S: x=(x_1, x_2)
\rightarrow (x_1, x_2, \Phi)$:
$$ \min_{\Phi} \int_\Om g(x)\sqrt{1+\Phi_{x_1}^2+\Phi_{x_2}^2}dx.$$
Our model shares a similar computational formulation, but is more self-contained in terms
of modeling and analysis since it is directly built on curves rather than on reference
points and surfaces. (We must point out that, in visual cognition, attention focuses do
play important roles in a number of cognitive tasks~\cite{kan}.)

Later in~\cite{illu_zhuchan}, Zhu and Chan proposed a more complex level-set based
variational model. Let $u:\Om\rightarrow\mR$ be an image defined on the domain $\Om$,
with $u=1$ on the objects and $u=-1$ otherwise. The key component of Zhu and Chan's model
is the {\em signed distance} function from solving the eikonal equation :
$$\frac{\partial d}{\partial t}=\text{sign}(u)(1-|\nabla d|), $$
with the initial data $d(x,t=0)=u(x)=u(x_1,x_2)$. Their variational model is then to
minimize the following energy for the level-set function $\phi$ which codes the desired
illusory contours:
\begin{equation} \label{E1:zhuchan}
\begin{split}
E(\phi)=&\int_\Om \left(
      (1+\mu C(d)\kappa^+(d))|d|\delta(\phi)\lvert\nabla \phi\rvert
          +\lambda H(d)H(\phi) \right) dx \\
&+\int_\Om\left( a+b\left\vert \nabla\cdot \left[ \frac{\nabla \phi}{\vert\nabla \phi
\vert} \right] \right\vert^2 \right) \vert\nabla \phi\vert\delta(\phi)dx,
\end{split}
\end{equation}
where $\delta$ denotes Dirac's delta, $H$ the Heaviside function, and $C$ a $C^1$
differentiable cut-off function. The $|d|\delta(\phi)\lvert\nabla \phi\rvert$ term
measures the distance from the zero level-set to the existent boundaries and
$C(d)\kappa^+(d)|d|\delta(\phi)\lvert\nabla \phi\rvert$ measures the distance to concave
corners. The second term of the first integral measures the overlap between the existent
objects and the interior of $\phi$. The last integral represents Euler's elastica energy
which was first proposed by Mumford~\cite{mum_elastica}, and later applied to geometric
image inpainting by Chan, Kang, and Shen~\cite{chakanshe}, and Esedoglu and
Shen~\cite{eseshe}. Zhu and Chan~\cite{illu_zhuchan} showed outstanding performance of
this model for several general examples. Our proposed model is much less complex than
this high-order geometric model, and is also independent of the signed distance function.
The essence of our proposed model is to be able to capture the most salient features of
illusory perception in a simple and analyzable manner, at the unsurprising cost of losing
certain high-order details. (Zhu and Chan~\cite{illu_zhuchan_shape} later also employed
prior shape information to capture illusory contours.)

%% working on the following afterwards *Jackie*

Mainly inspired by these pioneering works on quantitative modeling, we propose a new
energy-based model for studying illusory contours. The model is directly formulated on
admissible contours and does not depend upon reference points nor their associated
surfaces. The model has a much lower order of complexity which however yields good
leading-order approximation to most generic illusory contours (with exact capturing in
the canonical cases of Kanizsa's triangle or square). The relationship between this new
model and the more complex models of Zhu and Chan is formally analogous to that between
the first-order and second-order Taylor expansions in Calculus. Most remarkably, the
lower complexity of the new model allows to almost completely characterize the structures
of illusory contours, which provides a solid stepstone towards more general quantitative
analysis in the future. To our best knowledge, such rigorous characterization has been
unprecedented in the literature.

The second major feature of the new model is that in terms of computation, it can be
efficiently implemented by classical active contours and level-set based algorithms of
Osher and Sethian~\cite{oshset,osher_paragios,sethian_book,sethian_PNAS}. In particular,
we introduce a key but natural step of supervision for capturing robustly illusory
contours, which can also be considered as an innate component of the new model itself.

The paper has been organized as follows. In Section \ref{sec:2}, we develop and analyze
the new model for illusory contours that emerge from any generic configuration of real
{\em objects}. We are able to rigorously characterize the geometric structures of
illusory contours when identified as the local minima of the proposed model. In Section
\ref{sec:3}, the model is further extended to process configurations with real {\em
contours}. In Section \ref{sec:4}, an approximate or relaxed model is then developed
based on the classical active-contour model of Kass, Witkin and
Terzopoulos~\cite{kaswitter}, which allows the proposed model to be actually computed via
asymptotic approximation. In Section \ref{sec:5}, the level-set method of Osher and
Sethian~\cite{osher_paragios,oshset,sethian_book} is employed to compute the stable local
minima of the model, under the help of a key step of supervision.   Finally, the details
of numerical implementation as well as the computational performance of the proposed
model are provided in Section \ref{sec:6}.

%% Section 2.

\section{Illusory contours as stable local minima and characterization}\label{sec:2}

In this section, we propose a contour-based energy model, and model illusory contours as
the {\em local minima} of such an energy. The particular form of the energy allows
detailed and rigorous analysis of the geometric and topological structures of admissible
local minima. Such in-depth mathematical characterization is,  to our best knowledge,
unique in the literature.

Note that a local minimum corresponds to an energy valley and is stable {\em only} to
local perturbations. This differs our model from many other efforts in modeling illusory
contours as global minima. Aiming at local instead of global minima has also been
motivated by cognitive sensitivities of illusory perception (of light, geometry, or
objects, etc) to various factors including visual attention, focus points, orientations,
and scaling, etc. Such visual factors may disrupt or erase an observer's ongoing illusory
experience, while global minima often remain stable to these external perturbations.

In later sections we then develop an efficient level-set-based computational algorithm
that supports stable stagnation at such local minima.

\subsection{Kinks, outer and inner spans $\theta$, and a generic configuration $Q$}
 \label{sec:2.1}

Let $\Omega$ denote the entire open image domain, which could be the complete plane
$\mR^2$ or a finite rectangle area. Let $Q=\{ Q_i \mid i=1:k \}$ be a {\em configuration}
of {\em disjoint} compact domains whose boundaries are {\em piecewise}-$C^2$ simple
closed curves. Denote the boundaries by
 \[ \pd Q = \pd Q_1 \union \pd Q_2 \union \cdots \union \pd Q_k. \]

\begin{figure}[ht]
\centering{
  \epsfig{file=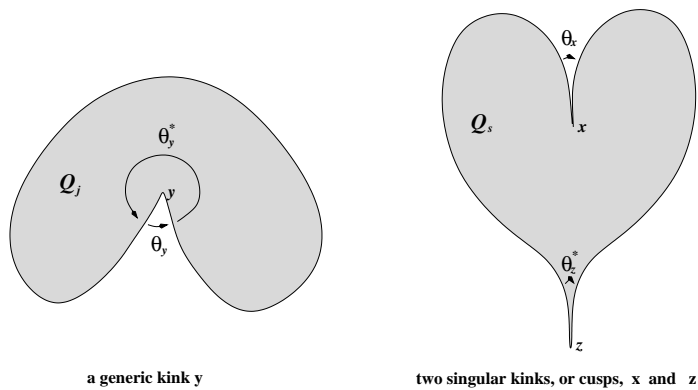,height=2in}
  \caption{{\bf Left panel:} a generic kink $y$ with outer span $\theta_y >0$
           and inner span $\theta^\ast_y=2\pi-\theta_y>0$. {\bf Right panel:} two examples of
           singular kinks or cusps; an inward cusp at $z$ with
           $\theta^\ast_z=0$ and an outward cusp $x$ with $\theta_x=0$. The current work
           discusses {\em generic} configurations containing only generic kinks.
           Perceptual effects of cusps have scarcely been experimented in visual cognition.}
  \label{F2:kinks}}
\end{figure}

\begin{defi}[A kink and its outer span $\theta$ and inner span $\theta^\ast$]
A point $y \in \pd Q$ is said to be a kink if the two tangent lines  from the two sides
are {\em not} identical. If $y$ is a kink, its outer span $\theta_y$ is the angle between
the two tangent lines when moving {\em continuously} from one arm to the other {\em
outside} $Q$ (see Fig.~\ref{F2:kinks}). We further define $\theta^\ast_y=2\pi - \theta_y$
to be the inner span. For a non-degenerate kink $y$, both $\theta_y$ and $\theta_y^\ast$
must be positive.
\end{defi}

A cusp is a singular kink point $y$ for which either the outer span $\theta_y=0$ or the
inner span $\theta_y^\ast=0$ (see Fig.~\ref{F2:kinks}). It will be  interesting in vision
psychology to understand the cognitive effects of cusps in various scales. In the current
work we shall assume that the given configuration $Q$ only contains {\em generic} kinks
so that $\theta_y, \theta^\ast_y >0$. Such a configuration $Q$ is said to be {\em
generic} consequently. Almost all classical examples correspond to generic
configurations.

\begin{defi}[Kink set $K$ and minimum spans $\theta_{\min{}}$ and $\theta^\ast_{\min{}}$]
Let $Q \subset \Omega$ be a generic configuration. Define its kink set $K[Q]$ to be the
collection of all kinks of $Q$, and its minimum outer span by
 \[ \theta_{\min{}} = \theta_{\min{}} [Q] =\min \{\theta_y \mid y \in K[Q]\}, \]
and minimum inner span by
 \[ \theta^\ast_{\min{}} = \theta^\ast_{\min{}}[Q]
    = \min \{ \theta_y^\ast \mid y \in K[Q] \}
    = \max \{ 2\pi -\theta_y \mid y \in K[Q] \}. \]
Since $\pd Q$ is compact and piecewise $C^2$, $K[Q]$ must be  finite and both
$\theta_{\min{}}$ and $\theta^\ast_{\min{}}$ are positive.
\end{defi}

%% above done on September 27. One xfig figure is needed

\subsection{Real and imaginary decomposition of a contour}

\begin{figure}[ht]
 \centering{
 \epsfig{file=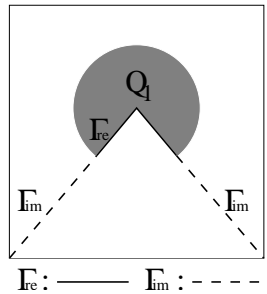, height=1.5in, width=1.36in}
 \caption{ Decomposition of $\Gamma$ into real and imaginary parts
           (partial view of Kanizsa's triangle).}
 \label{Fig21}  }
\end{figure}

Let $\Gamma$ be any piecewise $C^2$ simple closed curve in $\Omega$ {\em that does not
intersect the interior of $Q$}. Define its real and imaginary parts by
 \begin{equation}   \label{E2:reim}
  \Gamma_\re = \Gamma \joint \pd Q, \qquad \Gamma_\im = \Gamma \setminus \pd Q.
 \end{equation}
For an illusory contour $\Gamma$, the imaginary part corresponds precisely to the
illusory interpolation (Fig.~\ref{Fig21}). (The names ``real" and ``imaginary" have been
formally motivated by complex analysis in which a complex number $z$ can be decomposed
into real and imaginary parts: $z=z_\re +\sqrt{-1} \; z_\im$.)

Notice that since both $\Gamma$ and $\pd Q$ are compact and thus {\em closed} sets,
$\Gamma_\re$ must be closed in $\Omega$ and $\Gamma_\im$ {\em relatively} open in
$\Gamma$. We shall only consider a curve $\Gamma$ for which $\Gamma_\im$ contains {\em
finitely} many connected components to avoid unnecessary technicalities:
 \[ \Gamma_\im = \gamma^1_\im \union \gamma^2_\im \union \cdots \union \gamma^m_\im. \]
Since $\Gamma_\im = \Gamma \joint Q^c$, the intersection of a simple closed (and thus
locally connected) curve and an open domain, $\Gamma_\im$ must be locally
connected~\cite{kel97}. Consequently, each connected component $\gamma^j_\im$ must be a
connected curve segment with at most two endpoints.  {\em In the current work, we assume
that the two endpoints are distinct}. (Otherwise, the entire curve $\Gamma$ becomes a
loop which touches $\pd Q$ only at a single point. Such a loop can easily shrink to
decrease the energy to be proposed immediately next, and is hence irrelevant to the local
minima.) Let $\pd \gamma^j_\im$ denote the two boundary endpoints, and $\pd
\Gamma_\im=\pd \gamma_\im^1 \cup \cdots \cup \pd \gamma_\im^m$.

\subsection{Energy model and the structural theorem for local minima}

For a given generic configuration $Q$ in a 2-D open domain $\Omega$, define the class of
admissible curves
 \begin{equation}\label{E2:curves}
 \cC= \{ \Gamma \mid \mbox{$\Gamma$ is simple closed,
                           piecewise $C^2$ with finitely many pieces,
                           and $\Gamma \joint Q^\circ =\mathrm{empty}$}
      \}.
 \end{equation}
For any admissible curve $\Gamma \in \cC$, we define the energy
\begin{equation}\label{E1}
E_o[\Gamma] = E_o[\Gamma \mid \alpha, \beta]=\alpha\int_\Gr ds +\beta\int_\Gi ds,
\end{equation}
where $s$ denotes the arc-length parameter and $0<\alpha \ll \beta$. The subscript ``o"
in $E_o$ signifies that this energy is for {\em object}-based configuration $Q$. In the
next section, we will introduce an energy $E_c$ for {\em contour}-based configurations.
Notice that $E_o$ is the weighted sum of the lengths of $\Gr$ and $\Gi$. We are
interested in the behavior and characteristics of the local minimizers to $E_o$, in
particular, in how well they can quantify more familiar features of illusory contours in
visual cognition, which have often been subconsciously processed by human observers.

\begin{lemma}\label{L2:line}        %% Ln: means lemma in section n.
Let $\Gamma \in \cC$ be a local minimizer of $E_o$. Then each connected component of
$\Gi$ must be a straight line segment.
\end{lemma}

\begin{proof}
Suppose otherwise that one connected component $\gamma$ of $\Gi$ is not a straight line
segment. Then there must exist at least one point $b \in \gamma$, within {\em any} of
whose small neighborhoods, there can be found two points $a, c\in \gamma$ on the two
sides of $b$ such that $\{a, b, c\}$ are not collinear.  We then replace $\gamma|_{[a,
c]}$ by the straight line segment joining $a$ and $c$. Since $\Gamma$ and hence $\gamma$
are piecewise $C^2$ and $\gamma \subset Q^c$ (an open set), as long as the neighborhood
is small enough,  the straight line substitute does not intersect the rest of $\Gamma$
nor $Q$. Denote the new curve after such a micro-surgery by $\Gamma'$. Then $\Gamma' \in
\cC$ and $E_o[\Gamma'] < E_o[\Gamma]$ since the two only differ in between $a$ and $c$.
This contradicts to the assumption that $\Gamma$ is a local minimizer since such
micro-surgeries can occur in {\em any} local neighborhood of $b$.
\end{proof}

\begin{figure}[ht]
\centering{
  \epsfig{file=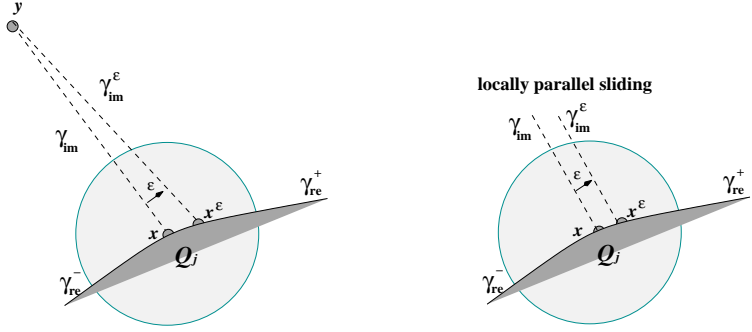,height=1.75in}
  \caption{Local perturbation at a junction point $x \in \pd \Gamma_\im \subset \pd Q$ can be
           approximated (in the first order of the displacement distance $\ep$) by
           a simple parallel translation of the attached imaginary ray $\gamma_\im$, due
           to Lemma~\ref{L2:line} and the same mechanism that makes direct solar rays
           appear approximately parallel when reaching onto the Earth. Parallel shifting
           simplifies most computation. }
  \label{F2:parallel}}
\end{figure}

In the following analysis, the local perturbation of a junction point $x \in \pd
\Gamma_\im$ along $\pd Q$ (see $x \to x^\ep$ in Fig.~\ref{F2:parallel}) has always been
accomplished by a simple parallel translation of the attached imaginary ray $\gamma_\im$.
This is justified by the preceding lemma, as well as the fact that when the other
endpoint $y$ is relatively far away, the perturbed imaginary ray $\overline{yx^\ep}$ due
to $x\to x^\ep$ is parallel to $\overline{yx}$ in the first order (of $\ep$) within small
neighborhoods of  $x$ (see Fig.~\ref{F2:parallel} and its caption).

\begin{lemma}[Two distinct imaginary rays cannot share a hinge] \label{L2:apart}
Suppose $\Gamma \in \cC$ is a local minimizer to $E_o=E_o[\cdot |\alpha, \beta]$, and
$\gamma^1$ and $\gamma^2$ are two distinct connected components (which are straight line
segments by the preceding lemma). Then there exists some critical ratio $r_c$, such that
as long as $r=\alpha/\beta < r_c$, $\pd \gamma^1 \joint \pd \gamma^2$ must be empty.
\end{lemma}

\begin{figure}[ht]
\centering{
  \epsfig{file=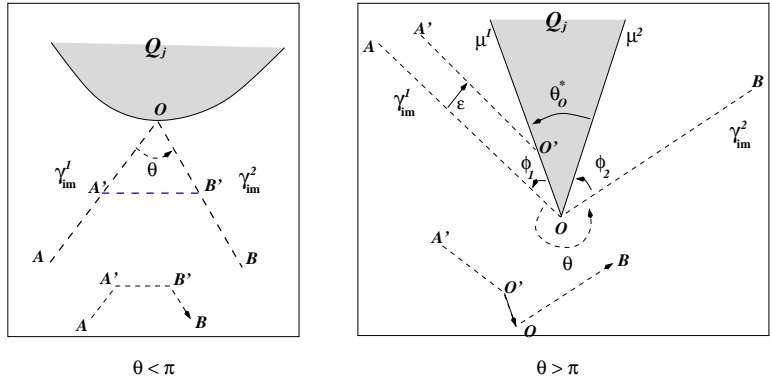,width=4in}
  \caption{Companion illustration for the proof of Lemma~\ref{L2:apart}:
           Two distinct imaginary rays $\gamma_\im^1$ and $\gamma_\im^2$
           (of a local minimum contour $\Gamma$) cannot share
            a common hinge $O$ (Lemma~\ref{L2:apart}).
           {\bf Left panel:} if $\theta <\pi$, a local perturbation of $\overline{AOB}$
           to $\overline{AA'B'B}$ decreases the total energy $E_o$.
           {\bf Right panel:} if $\theta \ge \pi$,
           a local perturbation of $\overline{AOB}$ to $\overline{A'O'OB}$ still decreases
           the total energy (see the computation in the proof of Lemma~\ref{L2:apart}). }
  \label{F2:hinge}}
\end{figure}

\begin{proof} (See the companion illustration in Fig.~\ref{F2:hinge})
Otherwise, assume that $\gamma^1$ and $\gamma^2$ share a hinge at a point $z$ (i.e., the
point $O$ in the figure). Then $z$ must be a {\em real} point, and one can assume that $z
\in \pd Q_j$ for some unique $j$.

Locally in a small diskette neighborhood of $z$, the two straight rays $\gamma^1$ and
$\gamma^2$ segment the disk into two parts: an ``inner" part that intersects with $Q_j$
and an ``outer" part. Let $\theta$ denote the outer span angle.

First suppose $\theta \in (0, \pi)$. Then any corner-cutting local deformation near $z$
(as shown in the left panel of Figure~\ref{F2:hinge}) can decrease energy $E_o$ since the
the real part of the energy is unchanged but the imaginary one shrinks. Thus this
scenario cannot occur.

Then one must have $\theta \ge \pi$. For $\pd Q_j$, similarly there are two {\em real}
tangent rays at $z$, denoted by $\mu^1$ and $\mu^2$, so that the acute cone of $\mu^1$
and $\gamma^1$ contains neither of the other two rays. Let $\phi_i$ denote the acute
angle between $\mu^i$ and $\gamma^i$, $i=1, 2$. Without loss of generality, assume that
$\phi_1 \le \phi_2$. Since
 \[
 \phi_1 + \phi_2 \le ( 2\pi - \theta) - \theta^{\ast}_{\min{}} \le \pi - \theta^{\ast}_{\min{}},
 \]
one has
 \[ \phi_1 =\min(\phi_1, \phi_2) \le \frac \pi 2 - \frac {\theta^{\ast}_{\min{}}} 2. \]
Then as shown in the right panel of Fig.~\ref{F2:hinge}, a small parallel sliding (with
perpendicular distance $\ep$) of the root of the imaginary ray $\gamma^1$ along the real
edge of $\pd Q_j$ in the direction of $\mu^1$ will introduce an energy increment of:
 \[
   \alpha \frac {\ep}{\sin \phi_1} - \beta \frac {\ep} {\tan \phi_1}
    = \frac {\ep \beta} {\sin \phi_1} \left(\frac \alpha \beta - \cos \phi_1  \right).
 \]
Define $r_c = \cos(\pi/2 - \theta^{\ast}_{\min{}}/2)=\sin (\theta^{\ast}_{\min{}}/2) >
0$. Then as long as $r=\alpha/\beta< r_c$, the leading order of energy increment is in
fact negative, which again contradicts to the given assumption. To conclude, as long as
$r < r_c$, two distinct imaginary line segments cannot share a hinge.
\end{proof}

\begin{figure}[ht]
\centering{
  \epsfig{file=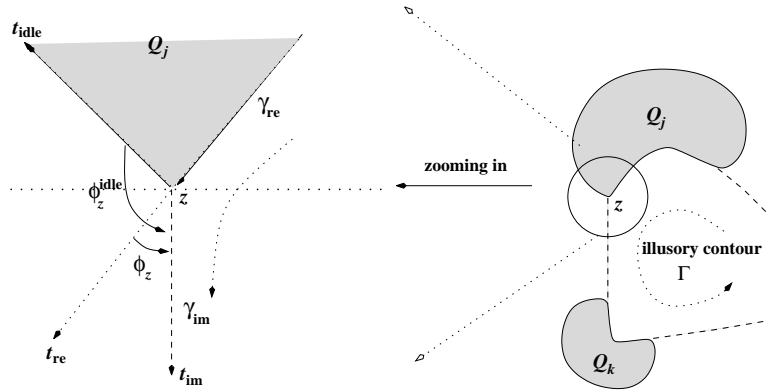,width=4in}
  \caption{A permissible geometric layout at a junction point $z \in J[\Gamma]$ with
           the turn $\phi_z < \pi/2$ and idle turn $\phi_z \ge \pi/2$, according to the
           results established in Lemmas~\ref{L2:turn} and~\ref{L2:idle}.  }
  \label{F2:junction}}
\end{figure}

 {
\begin{defi}[Junction set $J{[\Gamma]}$ and a turn $\phi_z$ at a junction point $z$]
\label{D2:turn}
 For a minimizer $\Gamma \in \cC$, we define its junction set to be
 \[
 J[\Gamma]  = \pd \Gamma_\im, \qquad \mbox{i.e., all endpoints of imaginary segments.}
 \]
By definition, one must have $J[\Gamma] \subseteq \pd Q$.  For any junction point $z \in
J[\Gamma]$, in one direction $\Gamma$ goes to a uniquely associated imaginary ray, and
let $\mt_\im =\mt_\im(z)$ denote this {\em outgoing} unit vector of orientation, and in
the other direction comes a curve segment of $\Gamma_\re$ (according to the preceding
lemma) and let $\mt_\re=\mt_\re(z)$ denote this {\em incoming} unit vector of orientation
(see Fig.~\ref{F2:junction}). The turn at $z$ is then defined to be the angle:
 \[ \phi_z = \arccos(\mt_\im \cdot \mt_\re) \in [0, \pi]. \]
\end{defi}
 }

\begin{lemma}[Acute turns at junction points] \label{L2:turn}
 Suppose $\Gamma \in \cC$ is a local minimum of $E_o$ and $z \in J[\Gamma]$ is a junction
 point. Then the turn $\phi_z < \pi/2$.
\end{lemma}

\begin{figure}[ht]
\centering{
  \epsfig{file=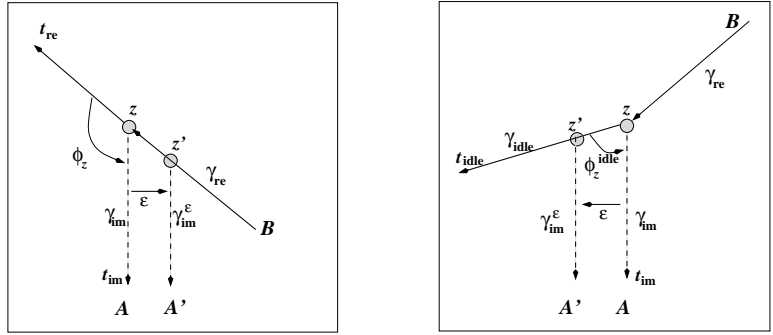,width=4in}
  \caption{Companion illustration for the proofs of Lemmas~\ref{L2:turn} and~\ref{L2:idle}.
           {\bf Left panel:} the turn $\phi_z$ at a junction point $z \in J[\Gamma]$ cannot be blunt,
           otherwise the local perturbation of $z$ to $z'$ and $\gamma_\im$ to $\gamma_\im^\ep$
           decreases both the real and imaginary energies in $E_o$. {\bf Right panel:} the idle
           turn $\phi^\idle_z$ cannot be acute for sufficiently small $\alpha/\beta$, otherwise
           the perturbation shown will cause a net drop in the total energy $E_o$ (see the
           computation in the proof of Lemma~\ref{L2:idle}). }
  \label{F2:turnanal}}
\end{figure}

\begin{proof}
Suppose otherwise that $\phi_z \ge \pi/2$. Then if one slides the root $z$ of the
imaginary ray $\gamma_\im$ along the {\em incoming} part of $\Gamma_\re$ while keeping
the imaginary rays parallel, as well as other parts of $\Gamma$ intact, as in the left
panel of Fig.~\ref{F2:turnanal}, the leading order energy increment becomes:
 \[
   -\beta \frac {\ep}{\tan(\pi - \phi_z)} - \alpha \frac {\ep}{ \sin(\pi-\phi_z) }
   < 0,
 \]
which contradicts to the assumption of $\Gamma$ being a  local minimum. (Notice that in
the critical case when $\phi_z=\pi/2$, the first term degenerates into a second-order
change $O(\ep^2)$. Thus the result still holds due to the second term.)
\end{proof}

This lemma has an intuitive interpretation in terms of visual cognition. Illusory
contours more or less are smooth and straight extrapolations of existing real
boundaries~\cite{chashe_cddinp,chashe_tvinp}, and consequently the turns at the junction
points should generally be as small as possible.

\begin{defi}[Maximum turn of a local-minimum contour $\Gamma$]
Suppose $\Gamma \in \cC$ is a local minimum of $E_o$. Define the maximum turn
$\phi_{\max{}}$ of $\Gamma$ by
 \[ \phi_{\max{}} = \max \{\phi_z \mid z \in J[\Gamma] \}.  \]
Then the preceding lemma implies that $\phi_{\max{}} < \pi/2$.
\end{defi}

\begin{defi}[Idle angle $\phi^\idle$ at a junction point]
Let $\Gamma \in \cC$ be a local minimum to $E_o$, and $z \in J[\Gamma]$ a junction point.
Define the imaginary tangent $\mt_\im=\mt_\im(z)$ as in Definition~\ref{D2:turn}. Further
define the {\em idle} tangent $\mt_\idle=\mt_\idle(z)$ to be the unit outgoing tangent
vector of $\pd Q \setminus \Gamma_\re$ (i.e., the {\em idle} real boundaries that do not
participate in the formation of $\Gamma$) at $z$ (see Fig.~\ref{F2:turnanal}). Define the
idle angle at $z$ by
 \[ \phi^\idle_z = \arccos( \mt_\idle \cdot \mt_\im ). \]
\end{defi}

\begin{lemma}[Behavior of idle angles at junctions] \label{L2:idle}
Suppose $\Gamma \in \cC$ is a local minimum to $E_o[\cdot \mid \alpha, \beta]$ for {\em
any} $\alpha, \beta > 0$, and $z \in J[\Gamma]$ is one of its junction point. Then
$\phi^\idle_z \ge \pi/2$.
\end{lemma}

\begin{proof}
Suppose otherwise that $\phi_z^\idle < \pi/2$. Then if one slides the root $z$ of the
imaginary ray $\gamma_\im$ along the {\em outgoing} idle part of $\pd Q \setminus
\Gamma_\re$ while keeping the imaginary rays parallel, as well as other parts of $\Gamma$
intact, as in the right panel of Fig.~\ref{F2:turnanal}, the leading order energy
increment becomes:
 \begin{equation} \label{E2:idlegle}
 \alpha \frac {\ep}{\sin \phi^\idle_z} - \beta \frac {\ep} {\tan \phi^\idle_z}
    = \frac {\ep \beta} {\sin \phi_z^\idle} \left(\frac \alpha \beta - \cos(\phi^\idle_z)  \right).
 \end{equation}
Since $\Gamma$ is a local minimum for {\em any} $\alpha, \beta >0$, one must have
 \[ - \cos(\phi^\idle_z) \ge 0, \qquad \mbox{or} \qquad \phi^\idle_z \ge \pi/2. \]
\end{proof}

Combining all the characterizations established in the lemmas of this section, we are
ready to state the main theorem of this paper.

\begin{thm}[Characterization of a local minimum] \label{T2:characterization}
 Let $Q$ be a generic configuration on an open domain $\Omega$, and $\theta_{\min{}},
 \theta^\ast_{\min{}} >0$ its minimum outer and inner spans. Suppose a given simple closed curve
 $\Gamma \in \cC$ satisfies the following structural conditions:
 \begin{enumerate}[(i)]
 \item (Imaginary behavior) each connected component $\gamma_\im$ of $\Gamma_\im$ is a
        straight line segment, and no two distinct components share a common hinge;
 \item (Junction behavior) at any junction point $z \in J[\Gamma]$, the turn $\phi_z <
         \pi/2$, and the idle angle $\phi^\idle_z \ge \pi/2$.
 \end{enumerate}
 Let $\phi_{\max{}}$ denote the maximum turn on $J[\Gamma]$. Then there exists a critical ratio
 $r_c=r_c(\theta_{\min{}}, \phi_{\max{}})< 1$, such that for any $\alpha$ and $\beta$ with
 $r=\alpha/\beta < r_c$, $\Gamma$ is a local minimum to the energy $E_o[\cdot \mid
 \alpha, \beta]$.
\end{thm}

\begin{proof}
We shall only outline the main steps of the proof and leave the details to the reader,
which are very similar to the proofs of the above ensemble of lemmas. Let $J[\Gamma]$
denote the junction set of such a given $\Gamma$, and define
 \[ \Gamma_\re^\circ=\Gamma_\re \setminus J[\Gamma]. \]
Then $\Gamma$ is segmented into three distinct parts:
 \[ \Gamma = \Gamma_\im \union \Gamma_\re^\circ \union J[\Gamma], \]
each of which leads to a distinct energy variation pattern due to local admissible
perturbations.

First, it is easy to see that under Condition (i) any local perturbations at a point of
$\Gamma_\im$ will increase the energy.

Next, $\Gamma_\re^\circ$ can be further partitioned into kink points $\Gamma_\re^\circ
\joint K[Q]$ and the majority rest of smooth points. At a smooth point $y$, any {\em
generic} local variation drifts a {\em real} local segment $\gamma$ into $Q^c$ and
converts it into an imaginary one, and hence introduces a net leading-order energy
increment of $(\beta-\alpha) \int_\gamma ds$, which is always positive for
$r=\alpha/\beta<1$. At a kink point $y$ on $\Gamma_\re^\circ$, one defines $r_1 = \sin
\theta_{\max{}}/2 > 0$. Then simple trigonometric calculus shows that for any $\alpha$
and $\beta$ with $\alpha/\beta < r_1$, any local admissible displacement of the kink
point always increases the energy.

Finally, one turns to the local perturbations at any junction point $z \in J[\Gamma]$. By
Condition (ii) on the acute turns, one must have that the maximum turn $\phi_{\max{}} <
\pi/2$. Define $r_2=\cos \phi_{\max{}} > 0$. Then trigonometric calculus shows that any
local drifting of $z$ into the {\em incoming} real contour $\Gamma_\re$ will increase the
energy as long as $\alpha/\beta < r_2$. On the other hand, any local drifting of $z$ into
the {\em idle} real edge $\pd Q \setminus \Gamma_\re$ always increases energy due to
Condition (ii) on $\phi^\idle_z \ge \pi/2$, for any $\alpha, \beta>0$. The details are
very similar to the proofs of Lemmas~\ref{L2:turn} and~\ref{L2:idle}.

In combination, one defines the critical ratio $r_c = \min (r_1, r_2, 1)$. Then for any
$\alpha$ and $\beta$ with $\alpha/\beta < r_c$, the above analysis in all the three
scenarios shows that the energy always increases under any local perturbations of
$\Gamma$. Thus $\Gamma$ must be a local minimum to $E_o[\cdot \mid \alpha, \beta]$.
\end{proof}

By this theorem, it is straightforward to establish the following corollary, which
demonstrates that local minima of $E_o$ are indeed reasonable models for illusory
contours, at least in terms of the leading-order approximation.

\begin{cor}
Both Kanizsa's triangle and square are local minimizers to $E_o[\Gamma \mid \alpha,
\beta]$ for sufficiently small $\alpha/\beta$.
\end{cor}

\begin{rmk}
In the preceding main theorem, only the condition $\phi^\idle_z \ge \pi/2$ could be
relaxed to $\phi^\idle_z \ge \pi/2-\delta$ for some constant $\delta=\delta(\alpha,
\beta)\ge 0$ with $\delta \to 0$ as $r=\alpha/\beta\to 0$ (due to
Eqn.~(\ref{E2:idlegle})). The preceding lemmas show that all the other conditions are
necessary and cannot be further relaxed.
\end{rmk}

%% above completed on September 30, 2005.
%% revised on October 5, 2005.

\section{Real contour bundles and object-contour mixtures} \label{sec:3}

In this section, we briefly discuss contour-based illusory contours, as well as those
arising from a mixture configuration of both objects and contours. Computationally,
however, it can be handled by exactly the same scheme for the object-based model, as to
be numerically demonstrated later on.

\begin{figure}[ht]
 \centering{
 \epsfig{file=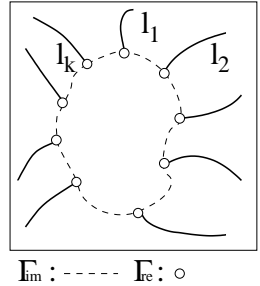, height=1.5in, width=1.31in}
 \caption{Decomposing a candidate contour $\Gamma$ into its real part $\Gamma_\re$
            and imaginary part $\Gamma_\im$.}
 \label{Fig22_1}  }
\end{figure}

Let $L=\{l_i\}^n_{i=1}$ be a bundle of disjoint compact contour segments on a given open
image domain $\Omega$. To be distinguished from illusory contours, each $l_i$ shall be
called a {\em real} contour (see Fig.~\ref{Fig22_1}). Let $\pd l_i=\{a_i, b_i\}$ denote
the two endpoints of $l_i$. Define
 \[
 \pd L = \union_{i=1}^n \pd l_i = \union_{i=1}^n \{a_i, b_i\}, \qquad
  L^\circ = L  \setminus \pd L.
 \]
As for object-based illusory contours, we define the class of contour-based admissible
contours by
 \[
   \cC_c
  =\{ \Gamma \mid
        \mbox{$\Gamma$ is simple closed, piecewise $C^2$, and $\Gamma \joint L^\circ=$empty}\}.
 \]
For each admissible contour $\Gamma\in \cC_c$, we define, in the same spirit of the
previous object-based case, the real-imaginary decomposition (Fig.~\ref{Fig22_1}):
 \[
        \Gamma_\re = \Gamma \joint \pd L, \quad \Gamma_\im
    =   \Gamma \setminus \pd L =\Gamma \setminus \Gamma_\re.
 \]
As a result, the previous object-based energy $E_o$ is modified to the following
contour-based:
 \begin{equation} \label{E3:Ec}
  E_c[\Gamma] = - \alpha \mathcal{H}^0(\Gr) + \beta \int_\Gi ds
              = - \alpha \mathcal{H}^0(\Gr) + \beta \int_\Gamma ds,
 \end{equation}
with $\alpha, \beta > 0$. Here $\mathcal{H}^0$ denotes the 0-dimensional Hausdorff
measure, or equivalently, the atomic counting measure.

The major difference of $E_c$ from the previous object-based energy $E_o$ resides in the
first negative term. Under this new formulation, any real end point from $\pd L$ acts
like an energy basin, in order to be able to anchor prospective illusory contours.

The local minimizers of $E_c$ are much easier to characterize.

\begin{thm}\label{T3:contourmini}
$\Gamma \in \cC_c$ is a local minimizer with a non-empty imaginary part $\Gamma_\im$ if
and only if it is a polygon whose vertices are all anchored within $\pL$.
\end{thm}

\begin{proof}
First suppose $\Gamma$ is a local minimizer with a non-empty imaginary part $\Gamma_\im$.
For any connected component $\gamma$ of $\Gamma_\im$, the same argument as in
Lemma~\ref{L2:line} applies since a local perturbation of $\gamma$ only modifies the
length integral in the $\beta$-term. Thus $\gamma$ must be a straight line segment. Since
$\gamma$ is arbitrary, $\Gamma \in \cC_c$ has to be a polygon whose vertices are anchored
within $\pL$.

Conversely, suppose $\Gamma \in \cC_c$ is a polygon whose vertices belong to $\pL$. Any
local variation {\em within} a side $\gamma$ of $\Gamma$ increases the $\beta$-term line
integral while leaving the atomic $\alpha$-term unchanged, since $\gamma$ is straight. On
the other hand, any nonzero local perturbation of a vertex of $\Gamma$ will drift it into
$\Omega \setminus L$ (since the perturbation must be kept within the admissible class
$\cC_c$), and consequently cause a leading-order energy increment of $\alpha$. Hence
$\Gamma$ is a local minimizer, and the proof is complete.
\end{proof}

\begin{figure}[ht]
 \centering{
 \epsfig{file=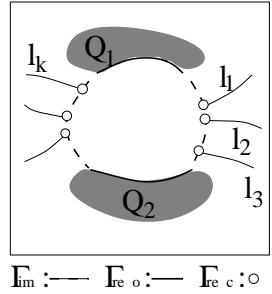, height=1.5in, width=1.37in}
 \caption{Decomposition of a candidate curve $\Gamma$ in an object-contour mixture configuration.}
 \label{Fig22_2}  }
\end{figure}

Finally, some examples in the literature also involve illusory contours from a mixture
configuration of both objects and contours. For this reason, we consider a mixture model
briefly. Consider an image which is composed of compact and mutually disjoint real
objects $Q$ and real contours $L$. Let $\Gamma$ be an admissible piecewise $C^2$ simple
closed curve in $\Omega$ that does not intersect $Q^\circ \union L^\circ$ (defined
previously). We define the real and imaginary part of $\Gamma$ as follows:
\begin{eqnarray*}
\Gamma_{\re} &=& \Gamma_{\text{re\_o}}\cup\Gamma_{\text{re\_c}}=(\Gamma \cap \PQ)\cup(\Gamma \cap \pL), \\
\Gamma_{\im} &=& \Gamma \setminus \Gamma_\re = \Gamma \setminus (\pd Q \union \pd L).
\end{eqnarray*}
Notice that here the real part has been further categorized into object-based real part
$\Gamma_{\text{re\_o}}$ and contour-based real part $\Gamma_{\text{re\_c}}$. We then
define a mixture energy functional $E_m$ in the form of:
\begin{equation*}
E_m[\Gamma]=   \alpha_o \int_{ \Gamma_{\text{re\_o}} } ds
             - \alpha_c \mathcal{H}^0(\Gamma_{\text{re\_c}})
            +  \beta\int_{\Gamma_\im} ds.
\end{equation*}
where $0<\alpha_o \ll \beta$ and $\alpha_c > 0$.

The characterization of a local minimizer to this mixture energy can be readily obtained
from a straightforward combination of the previous two major theorems. In particular, for
example, if $\Gamma$ is a local minimizer, then two distinct connected components
$\gamma^1$ and $\gamma^2$ of the imaginary part {\em can} share a common hinge, but the
hinge point must belong to $\pd L$ and not to $\pd Q$.

Next we discuss how to compute these models in real applications. In particular, we will
demonstrate numerically that for practical applications, a single relaxed  model is
sufficient to handle all the previous three theoretical models of $E_o, E_c,$ and $E_m$.

%% above done by September 30, 2005.

%% revised on October 6, 2005.

\section{Relaxation of the model: approximation by active contours}\label{sec:4}

The main models have been built upon the real-imaginary decomposition of any candidate
contour $\Gamma$, which in turn depends upon the identifiability of the {\em real} edge
set $\PQ$. Hence arise two major practical issues for  implementing these models
realistically. First, for general images, one has to perform a pre-processing of edge
detection in order to obtain $\PQ$, which is easier for binary images but highly
nontrivial for general continuous-tone images. Second, even provided with $\PQ$, it is
well known that direct computation of curves is highly challenging due to the densening
or coarsening effects of discrete points as well as potential topological changes like
splitting and merging.

The current section resolves the first issue, while the next section will address the
second.

The main result of this section is to show that  the edge-detection process can be
implicitly integrated into an approximate model, which surprisingly turns out to be the
more familiar active contour
model~\cite{chaves_ac,chashe_book,kaswitter,snake_KicKumOlver}. That is, the proposed new
model for illusory contours can be asymptotically approximated by active contours when
restricted in the class of admissible contours $\cC$.

First consider an ideal binary image $u(x)=\chi_Q(x)$, i.e., the characteristic function
of a generic configuration $Q$. Let $\eta(x)$ be any $C^\infty$ positive function which
is compactly supported on the unit disk in the plane, and for any scale $\sigma > 0$,
define the mollifier $\eta_\sigma(x)=\frac 1 {\sigma^2} \eta(\frac x \sigma)$, and the
mollification $u_\sigma=u \ast \eta_\sigma$. Since $Q$ is assumed to be compact in the
open image domain $\Omega$, $u_\sigma$ is well defined as long as $\sigma$ is smaller
than the distance $\mathrm{dist}(Q, \pd \Omega)$.

Then for any admissible curve $\Gamma \in \cC$, we introduce the following approximate
energy model
\begin{equation} \label{E4:ac}
E_a[\Gamma\vert u]=\alpha\int_\Gamma ds+\beta\int_\Gamma g(\lvert \nabla u_\sigma (s)
\rvert) ds,
\end{equation}
where $g(t)=\frac{1}{1+\lambda t^2}$ for some fixed parameter $\lambda>0$. This is in
fact precisely the classical active contour model, where $g(\lvert \nabla u_\sigma (s)
\rvert)$ is an edge signature distribution, or a {\em soft} edge
indicator~\cite{kaswitter}. The next theorem explains the relation between $E_o$ and
$E_a$.
\begin{thm} For any admissible curve $\Gamma \in \cC$, one has
$\displaystyle \int_\Gamma \frac{1}{1+\lambda\lvert\nabla u_\sigma
\rvert^2}ds\longrightarrow \int_{\Gi} ds$ as $\sigma\rightarrow 0$.
\end{thm}
\begin{proof}
Denote $B_\sigma =\lbrace x\in\Om \,\vert\,\text{dist} (x, \partial Q)\le \sigma\rbrace$.
Then $u_\sigma=1$ in $Q\setminus B_\sigma$ and $u_\sigma=0$ in $Q^c\setminus B_\sigma$.
As a result, $\nabla u_\sigma=0$ and $\frac{1}{1+\lambda\lvert\nabla u_\sigma
\rvert^2}=1$ in $\Om\setminus B_\sigma$. Since as $\sigma \to 0$, $B_\sigma$ shrinks to
$\PQ$, one has
\begin{equation} \label{QQ}
\frac{1}{1+\lambda\lvert\nabla u_\sigma \rvert^2} \longrightarrow 1 \text{ as }
\sigma\rightarrow 0 \text{ pointwise in } \Om\setminus\partial Q.
\end{equation}
Since $u$ is a characteristic function, one must have $\lvert \nabla u_\sigma\rvert
\rightarrow \infty$ as $\sigma\rightarrow 0$ along the piecewise smooth edge $\partial
Q$, and
\begin{equation}\label{QQQ}
\frac{1}{1+\lambda\lvert\nabla u_\sigma \rvert^2} \longrightarrow 0 \text{ as }
\sigma\rightarrow 0
\text{ on } \partial Q.
\end{equation}
In combination of \eqref{QQ} and \eqref{QQQ}, one has
$$\lim_{\sigma\rightarrow 0}{\frac{1}{1+\lambda\lvert\nabla u_\sigma \rvert^2}}
=\chi_{\Om\setminus\partial Q},$$ pointwise on the entire image domain $\Omega$. Since
$\Gamma \in \cC$ is piecewise smooth, simple closed, and $\Gamma \joint Q^\circ$ is
empty, by applying {\em Lebesgue's dominated convergence theorem} on $\Gamma$ with
respect to its arclength measure $ds|_\Gamma$~\cite{fol}, one has, as $\sigma \to 0$,
$$\int_\Gamma \frac{1}{1+\lambda\lvert\nabla u_\sigma \rvert^2}ds\longrightarrow
  \int_\Gamma \chi_{\Om\setminus\partial Q}ds=
  \int_{\Gamma \setminus Q} ds = \int_{\Gi}ds. $$
This completes the proof.
\end{proof}

Consequently, as $\sigma \to 0$, one has for any admissible contour $\Gamma \in \cC$,
 \[
 E_a[\Gamma \mid u] \longrightarrow \alpha \int_\Gr ds + (\alpha +\beta) \int_\Gi ds.
 \]
In particular, the current parameter pair $(\alpha, \alpha + \beta)$ corresponds to the
original parameter pair $(\alpha, \beta)$ in $E_o$. The approximate energy $E_a$ is,
however, computationally much more tractable since the practically difficult
decomposition $\Gamma=\Gr \union \Gi$ is replaced by much simpler function evaluations of
$g$ and $u_\sigma$. This technique is well known in the literature of active
contours~\cite{chashe_book,kaswitter}.

Furthermore, the active-contour approximation $E_a$, rigorously established above  for
binary images, is now readily applicable to general images which could have continuous
tones or have been degraded by moderate noises.

\begin{rmk}
Due to mollification, 1-D real contours in a mixture configuration are mollified to 2-D
stripes. As a result, our later computational results show that the relaxed
active-contour model works equally well for contour configurations or object-contour
mixtures.
\end{rmk}

\begin{rmk}
It is well known that classical (non-region-based) active contours are often easily
trapped in local minima~\cite{chashe_book}. This no longer becomes an issue herein since
illusory contours have been actually modelled as {\em local} minima, as extensively
developed in the preceding sections. That is, it would be bad news if the active contour
model $E_a$ never stably stagnates at local minima.
\end{rmk}

The key question becomes how to  compute robustly those local minima that correspond to
meaningful illusory contours in cognition. This issue is resolved by the {\em supervised}
level-set computation, which is the main theme of the next section.

%% above revised on Friday, September 30, 2005.
%% revised on Wednesday, Oct. 5, 2005.
%% revised on Oct. 10, 2005.

\section{Supervised level-set-based robust computation} \label{sec:5}

In this section, we derive the level set formulation to the approximate
model~\eqref{E4:ac} via classical active contours,  and introduce a simple but key step
of \textit{supervision} to be able to extract illusory contours as stable local minima.

\subsection{Derivation of the Euler-Lagrange equation}

The following brief derivation of the Euler-Lagrange equation is canonical (see,
e.g.,~\cite{chashe_book,kaswitter}) and makes the current paper more self-contained.

To derive the formal Euler-Lagrange equation, the candidate contour $\Gamma$ is assumed
to be at least $C^2$ so that second-order geometric features like curvature $\kappa$ are
well defined. We denote by $\Phi(x)=\Phi(x \mid u, \sigma)=g(\lvert \nabla u_\sigma (x)
\rvert)$ for convenience. Then
\begin{equation*}
E[\Gamma\vert u]=\alpha\int_\Gamma ds+\beta\int_\Gamma \Phi(s) ds,
\end{equation*}
 where $ds$ denotes the arc-length element of $\Gamma$.
 Since $\Gamma$ is assumed to be $C^2$, any smooth perturbation of $\Gamma_r$ of $\Gamma$
can be represented as a displacement along the normal $\mn$:
\begin{equation*}
\mc_r(s)=\mc(s)+r\delta \mc(s)=\mc(s)+r h(s)\mn(s), \,\,r\in (-\ep,\ep), \qquad h=O(1).
\end{equation*}
Then by Frenet's frame formula~\cite{chashe_book}, $\mn'(s)=- \kappa(s) \mt(s)$, and
$$\mc_r'(s)=(1-rh(s)\kappa(s))\mt(s)+rh'(s)\mn(s), $$
$$\lvert\mc_r'(s)\rvert=\sqrt{(1-rh(s)\kappa(s))^2+r^2h'(s)^2}.$$
Suppose $s\in [a, b]$ is the domain of $\mc$ and $\tilde{s}$ denotes the arc-length
parameter of $\mc_r$ so that $d\tilde{s}= |\mc_r'(s)|ds$. Then
$$\int_{\Gamma_r}d\tilde{s}=\int_a^b \lvert \mc_r'(s)\rvert ds.$$

\begin{equation*}
\frac{d}{dr}\int_{\Gamma_r}d\tilde{s}=\int_a^b \frac{\partial}{\partial r}\lvert\mc_r'(s)\rvert ds
= \int_a^b  \frac{-(1-rh(s)\kappa(s))h(s)\kappa(s)+rh'(s)^2}{\sqrt{(1-rh(s)\kappa(s))^2+r^2h'(s)^2}} ds.
\end{equation*}
\begin{equation} \label{R}
\begin{aligned}
\left.\frac{d}{dr}\, \right|_{r=0} \int_{\Gamma_r}d\tilde{s} &=\int_a^b -h(s)\kappa(s)ds\\
&= \int_\Gamma -\kappa(s)\mn(s)\cdot h(s)\mn(s)ds.
\end{aligned}
\end{equation}
 On the other hand, for the second term, by noticing that
 $$\frac{\partial}{\partial r} \Phi(\mc_r(s))=\nabla\Phi\cdot \frac{\partial}{\partial r}\mc_r(s)
    =\nabla\Phi\cdot h(s)\mn(s)=h(s)\frac{\partial\Phi}{\partial \mn}(\mc(s)), $$
 one has
 $$\frac{d}{dr}\int_{\Gamma_r} \Phi d\tilde{s}=
    \int_a^b \frac{\partial}{\partial r}\Phi(\mc_r(s))\lvert \mc_r'(s)\rvert ds
    +\int_a^b \Phi(\mc_r(s)) \frac{\partial}{\partial r} \lvert \mc_r'(s)\rvert ds, $$
\begin{equation} \label{RR}
 \left. \frac{d}{dr}\, \right|_{r=0} \int_{\Gamma_r} \Phi d\tilde{s}
=\int_a^b h(s)\frac{\partial\Phi}{\partial \mn}(\mc(s)) ds-\int_a^b
\Phi(\mc(s))h(s)\kappa(s)ds.
\end{equation}

In combination of \eqref{R} and \eqref{RR}, one has
\begin{equation*} %\label{QQQ}
\begin{aligned}
 \left. \frac{d}{dr} \, \right|_{r=0} E_a[\Gamma_r\vert u]
 =& -\alpha\int_\Gamma \kappa(s)\mn(s)\cdot \delta \mc(s) ds\\
  & -\beta \int_\Gamma \left(\Phi(s)\kappa(s)\mn(s)
    -\frac{\partial\Phi}{\partial\mn}(s)\mn(s)\right)\cdot \delta \mc (s) ds.
\end{aligned}
\end{equation*}
As a result, after introducing $t$ as the artificial evolution time, we arrive at the
gradient-descent equation for $E_a[\Gamma \vert u]$:
\begin{equation}\label{CE1}
\frac{\partial \mc}{\partial t}=(\alpha+\beta \Phi)\kappa\mn-\beta
%\frac{\partial \mc}{\partial t}=\left(\alpha+\beta g(\lvert \nabla u \rvert)\right)\kappa\mn-\beta
%\frac{\partial g(\lvert \nabla u_\sigma \rvert)}{\partial \mn}\mn
%\frac{\partial\Phi}{\partial \mn}\mn.
\Phi_\mn \mn.
\end{equation}
In particular, if one defines
 \[ G(x)=G(x\mid \alpha, \beta, \sigma, u)=\alpha +\beta\Phi(x)=
       \alpha +\beta g(\lvert\nabla u_\sigma(x)\rvert),
 \]
then \eqref{CE1} simplifies to
\begin{equation}\label{CE2}
\frac{\partial \mc}{\partial t}=G\kappa\mn-G_\mn \mn.
\end{equation}
This is the canonical ``snake" equation in its original form (see,
e.g.,~\cite{chashe_book,kaswitter}).

%% above revised by October 1, 2005

\subsection{Supervised computation via the level-set method}\label{super}

The level-set method of Osher and Sethian~\cite{oshset,osher_paragios,sethian_book} is a
powerful computational tool to handle curve and interface motions, especially those that
intrinsically involve topological changes. The above active contour equation has a simple
formulation under the level-set representation.

Suppose the evolving contour $\mc(t,s)$ is always identified as the zero-level contour of
an evolving level-set function $\phi(t, x)$, which is smooth or at least Lipschitz
continuous:
\begin{equation}\label{LS1}
\phi(t, \mc(t,s))=0, \; \forall t \ge 0.
\end{equation}
Assume for convenience that $\phi$ is positive in the interior of $\mc(t, s)$. By
differentiating (\ref{LS1}) with respect to $t$, one has
$$  \frac{\partial\phi}{\partial t}+\left\langle \nabla \phi,
    \frac{\partial \mc}{\partial t}\right\rangle=0.     $$
Recall that for an implicit contour, the curvature, and tangent and normal vectors can be
represented as follows:
\begin{equation}\label{Fo1}
\kappa=-\nabla\cdot\left[ \frac{\nabla \phi}{\lvert\nabla\phi\rvert}\right], \quad
    \mt=\frac{\nabla \phi^\bot}{\lvert\nabla\phi\rvert},    \quad
      \mn=\frac{\nabla \phi}{\lvert\nabla\phi\rvert}.
\end{equation}
Thus the active-contour model~\eqref{CE2} allows the following level-set representation:
$$\frac{\partial\phi}{\partial t}+\left\langle \nabla\phi,
-G\nabla\cdot\left[ \frac{\nabla \phi}{\lvert\nabla\phi\rvert}\right]\frac{\nabla
\phi}{\lvert\nabla\phi\rvert} -G_\mn\frac{\nabla
\phi}{\lvert\nabla\phi\rvert}\right\rangle=0.$$
 It can be further simplified to
\begin{equation}\label{Fo2}
\frac{\partial\phi}{\partial t}=
\lvert \nabla \phi \rvert G\nabla\cdot\left[ \frac{\nabla \phi}{\lvert\nabla\phi\rvert}\right]
+\lvert \nabla \phi\rvert G_\mn.
\end{equation}
Since $\nabla G= G_\mt \mt +G_\mn\mn$, one has
\begin{equation}\label{Fo3}
\nabla G\cdot \nabla \phi=\vert \nabla \phi \rvert (\nabla G \cdot \mn)=\vert \nabla \phi
\rvert G_\mn,
\end{equation}
and consequently,
\begin{equation}\label{Fo4}
\frac{\partial\phi}{\partial t}=
\lvert \nabla \phi \rvert
\nabla\cdot\left[ G\frac{\nabla \phi}{\lvert\nabla\phi\rvert}\right].
\end{equation}
In the variational-PDE literature, this evolutionary partial differential equation is
integrated via the Neumann boundary condition $\frac{\partial\phi}{\partial \nu}=0$ along
$\pd \Omega$.

%% above completed by Oct. 3, 2005.

Since in our model, illusory contours have been modelled as {\em local} minima, the
initial contour or the initial configuration of the level-set function becomes important.
In the numerical computation,  we have chosen the initial function $\phi_0$ such that
$\phi_0=1$ in $\Omega_\ep$ and $-1$ otherwise. Here $\Omega_\ep$ denotes an open
sub-domain of $\Omega$ such that the Hausdorff distance between $\Omega$ and $\Omega_\ep$
is less than $\ep$. That is, $\Omega_\ep$ is a slightly contracted version of $\Omega$.
As long as the targeted illusory contour is a compact set on $\Omega$, $\Omega_\ep$ can
capture it within for small enough $\ep$.

Finally, unlike the classical active-contour computation whose goal is to capture {\em
real} boundaries of existent objects, the active-contour model $E_a$ in the present
context approximates our original contour model $E_o$ and hence is targeted at capturing
illusory contours. There is a side condition which has been stated theoretically
throughout the earlier discussion but not yet incorporated into the level-set
formulation. In the definition of the admissible class $\cC$ of candidate illusory
contours, we have required that any admissible contour $\Gamma$ must lie {\em outside}
the object configuration $Q$, i.e., that $\Gamma \joint Q^\circ$ is empty.

Classical level-set functions for active contours ideally prefer $\phi>0$ on $Q$ in order
to capture  {\em real} edges and {\em real} objects, but the above discussion shows that
for the computing of illusory contours, ideally one prefers $\phi < 0$ on the object
configuration $Q$. Thus we impose the following supervision condition for the level-set
computation:
 \[ \phi(t, x) \equiv -d_Q(x), \qquad x \in Q^\circ, \;\; t>0, \]
where $d_Q(x) =\mathrm{dist}(x, \pd Q)$ is the absolute distance function. In reality,
for computation on a discrete image lattice, the supervision can be conveniently replaced
by the less complex one:
 \[ \phi(t, x) \equiv -1, \qquad x \in Q^\circ, \;\; t>0 . \]
On one hand, it substantially lessens the computational burden without turning to the
eikonal equation for $d_Q(x)$, and on the other, it provides a cheaper way to minimizing
the second term in Zhu and Chan's model~\eqref{E1:zhuchan} without turning to either the
variational formulation or its associated Euler-Lagrange equation. To conclude, the last
supervision condition is an efficient tool to reinforce the contour admissibility
condition ``$\Gamma \joint Q^\circ=\mbox{empty}$" in the original exact model.

%% above revision is completed on October 3, 2005.

\section{Implementation details and computational examples}\label{sec:6}

In this last section, we explain the numerical implementation and discuss several generic
numerical examples. The equation and conditions in the preceding section \ref{super} can
be reiterated by
\begin{equation}\label{leqn}
\left\{\begin{aligned}
&\frac{\partial\phi}{\partial t}=\lvert \nabla \phi \rvert
\nabla\cdot\left[ G\frac{\nabla \phi}{\lvert\nabla\phi\rvert}\right]
 \text{ for } (t,x)\in (0, \infty)\times \Om,\\
&\phi(0,x)=\phi_0(x), \\
&\frac{\partial\phi}{\partial \nu}=0 \text{ for } (t,x)\in (0,\infty)\times\partial\Om, \\
&\phi(t,x)=-1 \text{ for } (t,x)\in (0,\infty)\times Q^\circ,
\end{aligned}
\right.
\end{equation}
where $G(x)= \alpha +\beta g(\lvert\nabla u_\sigma(x)\rvert)$.

Let $\phi^m_{i,j}$ be the approximation to the value $\phi(t_m,x_i^1,x_j^2)$, where
$x_i^1=i\Delta x_1$, $x_j^2=j\Delta x_2$ with $x=(x_1, x_2)$, and $t_m=m\Delta t$. For
simplicity, we assume the spacial sizes coincide: $\Delta x_1=\Delta x_2 =1$. Then,
\begin{equation}\label{level1}
\phi^{m+1}_{i,j}\approx\phi^m_{i,j}+\Delta t\lvert\nabla \phi^m_{i,j}\rvert \nabla\cdot
\left[\left(\alpha+\beta g(\lvert(\nabla u_\sigma)^m_{i,j}\rvert)\right) \frac{\nabla
\phi^m_{i,j}}{\vert\nabla \phi^m_{i,j}\vert_\epsilon}\right],
\end{equation}
where $|\cdot |_\epsilon$ is defined by
$$|s|_\epsilon:=\sqrt{s^2+\epsilon^2},$$
for some positive $\epsilon \ll 1$. This simple regularization technique helps the
computation of the denominator $|\nabla \phi|$ when it is close to be singular~(see,
e.g.,~\cite{rudoshfat,she_dejitter,she_weber}). We now focus on the spatial
discretization of the right hand side at a fixed time $t_m$, and shall leave out the
superscript $m$ for convenience. We have employed central differencing for differential
discretization, and mean filtering for interpolation. Write
$$\overrightarrow{V}_{i,j}=(V^1_{i,j},V^2_{i,j})=
\left(\alpha +\beta g(\vert (\nabla u_\sigma)_{i,j}\vert \right)) \frac{\nabla
\phi_{i,j}}{\vert\nabla \phi_{i,j}\vert}.$$
 Based on central differencing at the half points of the Cartesian lattice, one has
\begin{equation}\label{dis1}
\begin{aligned}
\nabla\cdot\overrightarrow{V}_{i,j}&=\frac{\partial}{\partial x}V^1_{i,j}+\frac{\partial}{\partial y}V^2_{i,j}\\
&\approx\left(V^1_{i+\frac{1}{2},j}-V^1_{i-\frac{1}{2},j}\right)+\left(V^2_{i,j+\frac{1}{2}}-V^2_{i,j-\frac{1}{2}}\right).
\end{aligned}
\end{equation}
Thus we need to specify the half-point values in \eqref{dis1}. For example, at
$(i+\frac{1}{2},j)$, one has
\begin{equation*}
\begin{aligned}
\frac{\partial}{\partial x}\phi_{i+\frac{1}{2},j}&\approx\phi_{i+1,j}-\phi_{i,j},\\
\frac{\partial}{\partial y}\phi_{i+\frac{1}{2},j}&\approx
\frac{1}{2}\left( \frac{\phi_{i+1,j+1}-\phi_{i+1,j-1}}{2}+\frac{\phi_{i,j+1}-\phi_{i,j-1}}{2}\right).
\end{aligned}
\end{equation*}
 To satisfy the supervision condition $\phi(t,x)=-1$ for $(t,x)\in
(0,\infty)\times Q^\circ$, we reset $\phi^{m+1}_{i,j}=-1$ on $Q^\circ$ after each
discrete evolution~\eqref{level1}. For more details on numerical discretization, we also
refer the reader to~\cite{chaoshshe}.

%% above revision is done by Oct. 3, 2005.

\begin{figure}[ht]
 \centering{
 \epsfig{file=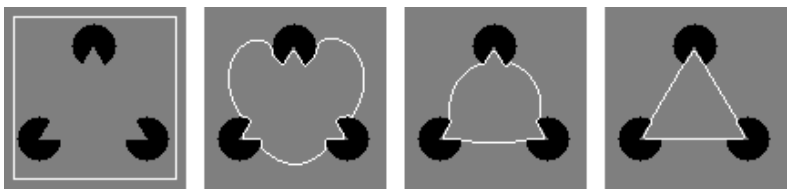, height=1in, width=4.14in}
 \caption{Kanizsa' illusory triangle as a stable local minimum to the proposed contour energy.}
 \label{Fig13}  }
\end{figure}

\begin{figure}[ht]
 \centering{
 \epsfig{file=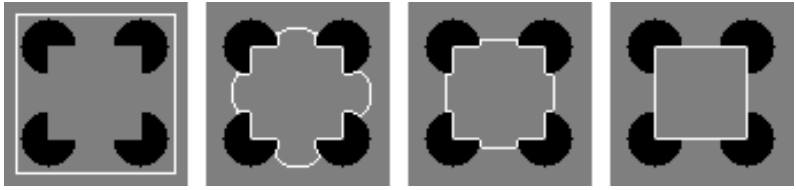, height=1in, width=4.17in}
 \caption{Kanizsa' illusory square as a stable local minimum and the stagnation process.}
 \label{Fig14}  }
\end{figure}

\begin{figure}[ht]
 \centering{
 \epsfig{file=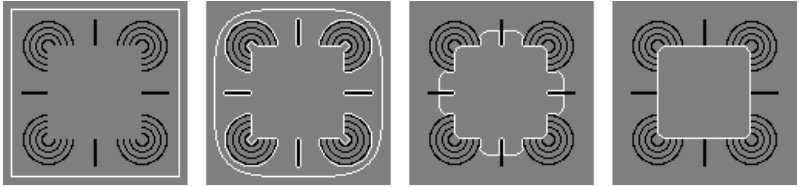, height=1in, width=4.19in}
 \caption{An illusory square emerges from a configuration of {\em real}
          contour bundles. The {\em object}-based supervised level-set computation works equally well
          for {\em contour}-based illusory contours due to the mollification operator applied to a given
          image $u$. }
 \label{Fig15}  }
\end{figure}

\begin{figure}[ht]
 \centering{
 \epsfig{file=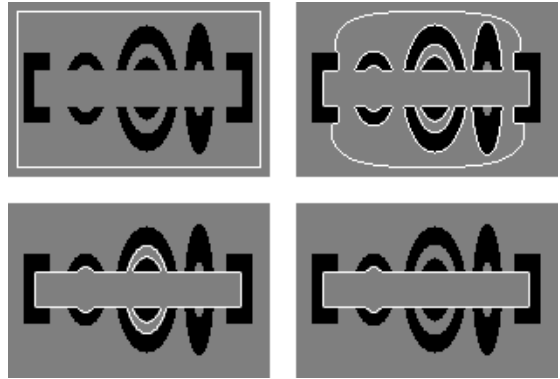, height=2in, width=2.91in}
 \caption{An illusory contour emerges from a complex configuration $Q$ of objects,
          as captured by the supervised level-set-based active contours. The
          small hump in the final output (near the left end of the illusory bar) is not
          due to numerical defection but to the competition between the two weights
          $\alpha$ and $\beta$ in the model. }
 \label{Fig16}  }
\end{figure}

\begin{figure}[ht]
 \centering{
 \epsfig{file=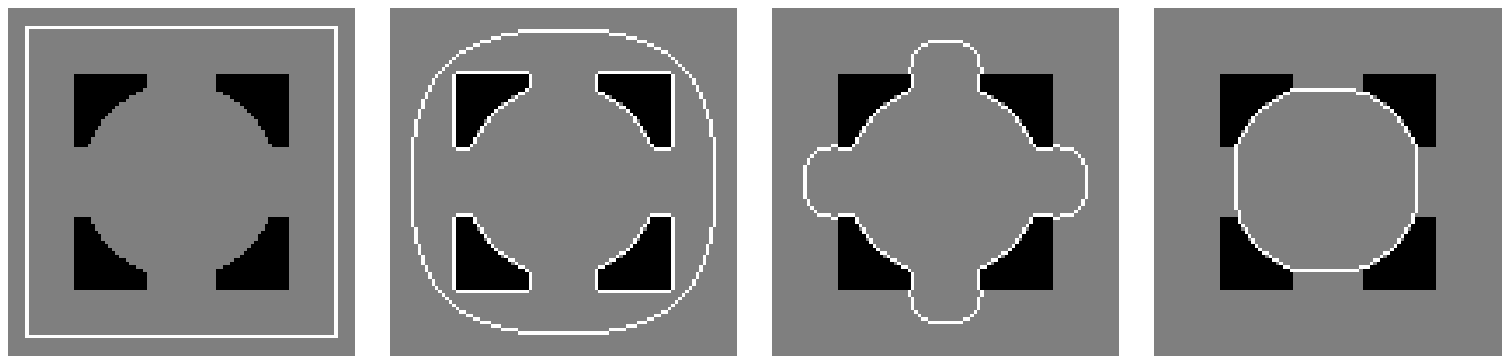, height=1in, width=4.18in}
 \caption{The consequence of first-order geometric modeling: our model can efficiently capture
          illusory objects and contours in their first-order proximity. To achieve high order
          fidelity, the model should incorporate at least the curvature information,
           as in the earlier works of Zhu and Chan~\cite{illu_zhuchan,illu_zhuchan_shape}.
           Such high order models are, however,
           often highly challenging for both rigorous theoretical analysis and efficient
           numerical computation. In-depth analysis in Section~2 is often
           difficult to establish for high order models. }
 \label{Fig17}  }
\end{figure}

Shown in Fig.~\ref{Fig13} through~\ref{Fig17} are computational outputs on some generic
examples of illusory contours, including the well known classical examples of Kanizsa's
triangle (Fig.~\ref{Fig13}) and square (Fig.~\ref{Fig14}).

Fig.~\ref{Fig15} shows an example of illusory contours from a configuration of contour
bundles, as theoretically modelled in Section~\ref{sec:3}. Due to the mollification
effect, the supervised level-set-based active contour model, which has been mainly
developed for {\em object}-based illusory contours, works equally well for {\em
contour}-based illusory contours in terms of computation.

Fig.~\ref{Fig16} shows a computed illusory contour emerging from a more complex
configuration $Q$ of objects. The small hump remaining in the final output is not due to
numerical defection but to the competition between the two terms in the model $E_o$, or
equivalently, between $\alpha$ and $\beta$.

Finally, Fig.~\ref{Fig17} shows the first-order nature of our proposed model. In order to
be able to output smoother illusory shapes, high order geometric features such as the
curvature have to be incorporated, as in Zhu and Chan's earlier
works~\cite{illu_zhuchan,illu_zhuchan_shape}. The tradeoff is that comprehensive
analytical results as in Section~\ref{sec:2} would be extremely difficult to develop,
which is well known in the literature of variational modelling involving
curvatures~\cite{chakanshe,mum_elastica}.

The captions of the figures also provide further on-site explanation and discussion.

\section*{Acknowledgments}

The authors would like to acknowledge the intellectual benefits from the Institute of
Mathematics and its Applications (IMA) for her constant support to imaging and vision
sciences. Shen also would like to thank Professors Alan Yuille in the Statistics
Department at UCLA and Dan Kersten in the Psychology Department at the University of
Minnesota for their inspiration and support to his research on cognitive vision.

%\bibliographystyle{plain}
%\bibliography{summer}
%% the following bibitems have been generated automatically from the above two lines

\end{document}